%% file: main.tex
\documentclass{article}

\usepackage[final]{neurips_2026_vericode}
\usepackage[T1]{fontenc}
\usepackage{hyperref}
\usepackage{url}
\usepackage{booktabs}
\usepackage{amsmath}
\usepackage{amsfonts}
\usepackage{microtype}
\usepackage{xcolor}
\usepackage{graphicx} \usepackage{listings}
\usepackage{colortbl}

\newcommand{\ours}{\textsc{FormalWeaver}}
\newcommand{\artifact}{\textsc{FloatSpec}}

\title{\ours{}: Dependency-Aware Translation of Formal Repositories at Scale}
\workshoptitle{NeurIPS 2026 Workshop on AI for Verifiable Coding}

\author{%
  \textbf{Hantao Lou}\textsuperscript{1,\ensuremath{\dagger}}%
  \thanks{%
    \textsuperscript{1}University of California, Berkeley
    \textsuperscript{2}Independent Researcher
    \textsuperscript{\ensuremath{\dagger}}Corresponding author.
    Author Emails: \{hantaolou.htlou, alokbeniwal\}@gmail.com.
  },
  \textbf{Alok Singh}\textsuperscript{2}
}

\begin{document}

\maketitle

\begin{abstract}
AI agents are increasingly capable of generating formal code and proofs under given specifications, but repository construction requires coherence of specifications and proofs across dependencies, beyond isolated theorem completion. Existing evaluations largely focus on declaration-level success and therefore do not directly measure whether an agent can preserve source coverage, specification semantics, and proof correctness together while growing an integrated repository. We study this gap through \textit{formal repository translation}, where the task is to translate formal source repositories into another formal language. We introduce \ours{}, a source-grounded pipeline that translates dependency-ready items against frozen upstream representations and interfaces, with compiler-guided repair scoped to the current item. Applied to Flocq, a Coq library for floating-point reasoning, the pipeline produces \artifact{}, a Lean port with 58 core files, 5,075 indexed items, and 194,704 lines of Lean code. We further conduct controlled experiments comparing \ours{} with different translation strategies on submodules of the Flocq translation task. We evaluate two aspects in sequence: whether a translation compiles and, for those that compile, whether it is semantically aligned with the source. Their conjunction defines translation accuracy. The results show that \ours{} substantially improves translation accuracy over simpler translation strategies.
\end{abstract}

\section{Introduction}

Formal-generation systems based on large language models (LLMs) are usually evaluated one declaration at a time. A model receives a theorem, function, or specification and succeeds when the target prover accepts the result~\citep{yang2023leandojo,ye2026verina,thakur2026clever}. In a repository, declarations are connected: a representation chosen in a foundational module constrains downstream specifications and proofs, and a local repair can invalidate previously accepted work. Scaling formal generation to repositories therefore introduces a coherence problem across dependencies and interfaces~\citep{xin2026verisoftbench,li2026p,ye2026vero}.

\begin{figure*}[t] \centering \includegraphics[width=\textwidth]{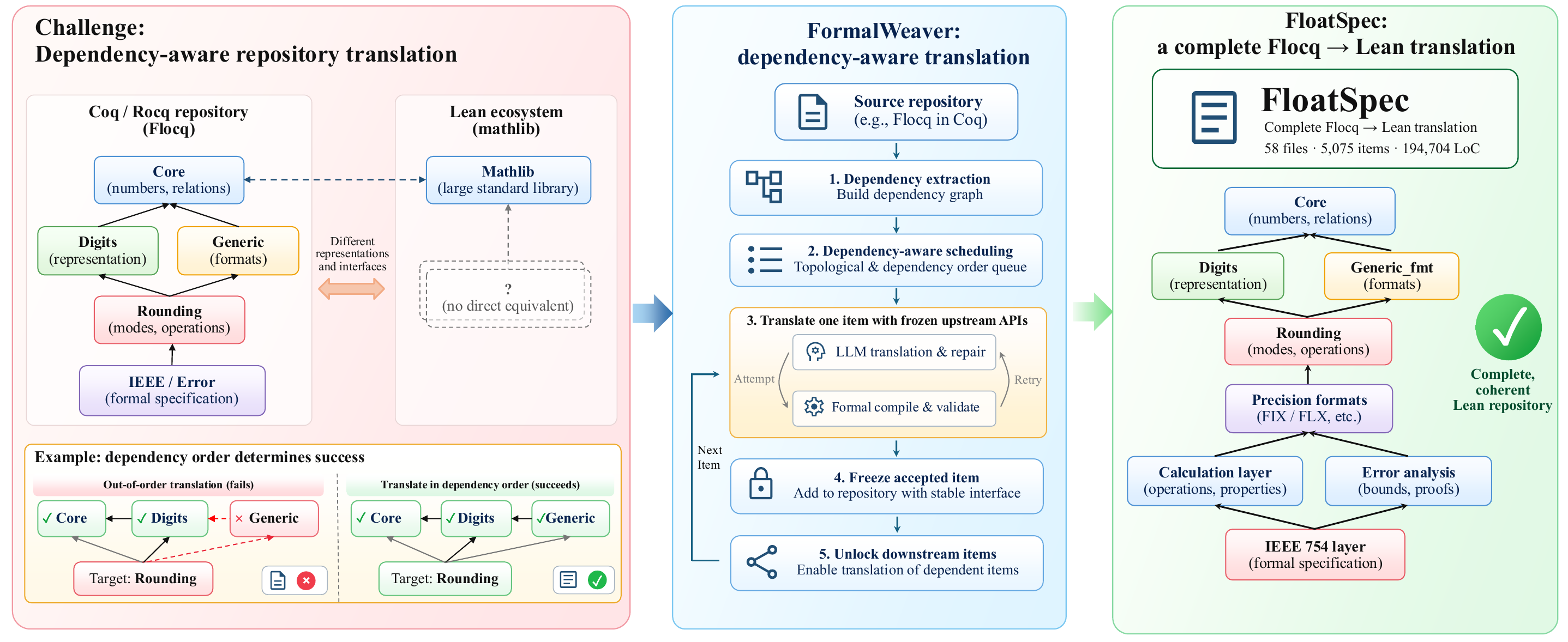} \caption{Overview of our main challenge, method, and contribution. \textbf{Left:} Cross-prover repository translation must preserve dependency order despite representation and interface mismatches. \textbf{Center:} \ours{} extracts dependencies, schedules ready items, compiles and repairs translations against frozen upstream APIs, and unlocks downstream work only after acceptance. \textbf{Right:} Applied to Flocq, \ours{} produces \artifact{}, a complete and coherent Lean repository with 58 files, 5,075 indexed items, and 194,704 lines of code.} \label{fig:pipeline} \end{figure*}

Formal knowledge is also fragmented across proof-assistant ecosystems. Mature libraries encode substantial formalization effort, yet their declarations and proofs generally cannot be reused directly in another prover~\citep{stoskopf2025babelformal,wu2026itpeval}. \emph{Formal repository translation} offers a way to transfer this knowledge: the source supplies explicit declarations, dependencies, and semantics against which the target development can be evaluated~\citep{viennot2025minif2f,stoskopf2025babelformal,wu2026itpeval}.

Translation complements proof synthesis, which starts from a target-language specification, and autoformalization, which recovers formal statements from informal descriptions~\citep{wu2022autoformalization,azerbayev2023proofnet,poiroux2025reliable}. A formal source makes the intended statements explicit, but adapting them to another ecosystem still requires compatible representations, interfaces, and proofs. Repository-context benchmarks show that proof success degrades as transitive dependencies grow, while cross-prover evaluation identifies library and ecosystem mismatch as a major bottleneck~\citep{xin2026verisoftbench,wu2026itpeval}. These observations motivate our central research question:

\begin{quote}
\emph{Can LLM agents efficiently translate complex formal repositories while preserving source semantics and coherence across dependencies and interfaces?}
\end{quote}

To address this question, we introduce \ours{}, a source-grounded pipeline for dependency-aware formal repository translation. \ours{} indexes source declarations and their dependencies, decomposes repository translation into item-level tasks executed in dependency order, stabilizes upstream representations and interfaces before downstream generation, and scopes compiler-guided repair to the current task to prevent local fixes from silently rewriting established dependencies. Using \ours{}, we construct \artifact{}, a complete Lean translation of Flocq~\citep{boldo2011flocq,boldo2015verified}, a mature Coq library for IEEE~754 floating-point reasoning. We further conduct controlled experiments on dependency-connected subsets of the Flocq translation task, comparing \ours{} against flat, item-wise, and compiler-guided translation strategies under matched settings. We measure translation accuracy through a compilation step followed by source-grounded semantic alignment: an item is counted as accurate only if it passes both checks. Our results show that the repository-level coordination introduced by \ours{} substantially improves translation accuracy and the ability to construct an integrated formal repository.

Our contributions are:

\begin{enumerate}
\item \textbf{A dependency-aware approach to formal repository translation.}
We identify repository-level coherence across dependencies, interfaces, and semantics as a central challenge in scaling formal translation, and introduce \ours{}, which combines dependency-ordered translation, stabilized upstream interfaces, scoped compiler-guided repair, and source-grounded validation.

\item \textbf{\artifact{}, a large-scale Lean translation of Flocq.}\footnote{\artifact{} is publicly available at \url{https://github.com/Beneficial-AI-Foundation/FloatSpec}.}
Using \ours{}, we construct a complete Lean port of Flocq with 58 core files, 5,075 indexed items, and 194,704 lines of Lean code, demonstrating the feasibility of translating a substantial formal repository across proof ecosystems.

\item \textbf{A controlled study of repository-scale translation strategies.}
We compare \ours{} with flat, item-wise, and compiler-guided translation strategies. We evaluate compilation first and then assess semantic alignment for compiling translations, aggregating the two checks into translation accuracy. Our results show substantial advantages from dependency-aware repository construction.
\end{enumerate}

\section{\ours{} for Formal Repository Translation}

We study how to translate a dependency-rich formal repository from one proof assistant to another using LLM agents. Unlike item-level translation, repository translation requires translated declarations to coexist under a shared set of representations, interfaces, and dependencies. We introduce \ours{}, which converts this repository-level problem into dependency-aware item translation while maintaining a consistent target repository throughout construction.

\subsection{Problem Formulation}

We represent the source repository as a dependency graph. Each node is a
source declaration, such as a definition or theorem; an edge from one declaration
to another means that the second depends on the first. Translation constructs a
target repository and records the correspondence between source declarations and
their target counterparts.

We evaluate each translated item in two steps. First, the target
declaration must compile in the accumulated repository. Second, a compiling
declaration must be semantically aligned with its source counterpart rather
than merely type-checking in the target system. This second check is necessary
because target-side acceptance alone does not guarantee semantic equivalence
across proof assistants~\citep{wu2026itpeval}. We count an item as an accurate
translation only when it passes both checks.

At the repository level, \emph{coverage} records which in-scope source
declarations have target counterparts. \emph{Integration} records whether
accepted declarations work together under shared representations, interfaces,
and dependencies. These properties describe the repository in which the two
evaluation steps are performed; they are not additional components of the
translation-accuracy metric.

\subsection{Dependency-Aware Translation Pipeline}

Dependencies create two distinct challenges: translating an item requires its
prerequisites to be available, while changing an upstream interface can
invalidate downstream work. Dependency-ready scheduling addresses the first;
stabilized interfaces and scoped repair address the second. The center panel
of Figure~\ref{fig:pipeline} summarizes the \ours{} pipeline: repository
planning, dependency-ready translation and repair, and validation and integration.

\paragraph{Repository planning.}
\ours{} first indexes the source repository using prover metadata, extracting
declarations, module membership, and dependency edges. Each source declaration
is assigned a stable identifier and target location. The resulting dependency
graph determines which items are ready for translation and provides an explicit
source-target inventory for tracking coverage.

Repository planning also establishes target-side representations and public
interfaces before downstream use, since proof ecosystems differ in libraries,
primitive operations, and representation conventions~\citep{stoskopf2025babelformal,wu2026itpeval}.
For example, using mathematical reals or machine floating-point values changes
the meaning of later rounding specifications. Freezing such choices makes their
downstream consequences explicit before proof code accumulates.

\paragraph{Dependency-ready translation.}
An item is scheduled only after the target counterparts of its required
predecessors are available. The worker translates that item against their
established signatures; after acceptance, its public interface is frozen and
made available to dependents. Revising an accepted interface is therefore an
explicit repository event that invalidates all items that depend on it,
directly or indirectly, rather than an incidental side effect of repairing a later item.

\paragraph{Item translation and repair.}
For each dependency-ready item, the worker receives the source declaration,
relevant local documentation, and the established target signatures of its
dependencies. The worker generates the corresponding Lean declaration and
invokes the target compiler. Compilation errors are returned to the same worker
for bounded repair.

Repair is scoped to the current item; accepted upstream declarations remain
read-only. After local compilation succeeds, the item is checked in the
accumulated target environment before it can unlock downstream work.

\paragraph{Validation and integration.}
Evaluation follows two sequential steps. The compilation step first verifies
that generated code is valid Lean in the accumulated target repository. Only
translations that pass this step proceed to semantic-alignment evaluation.
Depending on the item, semantic alignment may use independently checked
source-target correspondence, bilateral native execution under an explicit
representation relation, or additional source-grounded review. Cases without
sufficient evidence are not counted as accurate solely because they compile.

An item ledger records the source-target mapping, compilation state, semantic
evidence, and dependency state. Section~4 aggregates the two evaluation aspects
into translation accuracy: an item succeeds only when it compiles and is
semantically aligned with its source.

\section{\artifact{}: A Complete Lean Translation of Flocq}

We apply \ours{} to Flocq~\citep{boldo2011flocq,boldo2015verified}, a mature
Coq library for formal reasoning about floating-point arithmetic. The right
panel of Figure~\ref{fig:pipeline} summarizes the resulting \artifact{}
translation. This section describes its scope and architecture, the released
artifact, and a representative repository-level translation challenge.

\subsection{Scope and Architecture}

Flocq provides radix-parametric definitions and theorems for floating- and fixed-point arithmetic, including IEEE~754 arithmetic, and has served as infrastructure for verified floating-point computation~\citep{boldo2011flocq,boldo2015verified}. The source development contains 40 files, 2,164 items, and 24,786 lines of Coq code. \artifact{} ports this development to Lean~4 and Mathlib~\citep{demoura2021lean4,mathlib2020}. The target contains 58 core files, 5,075 indexed items, and 194,704 lines of Lean code. Given the scale of the resulting repository, we evaluate the semantic faithfulness of the translation using the repository-scale semantic validation framework introduced in concurrent work~\citep{anonymous_transformal}.

The translated repository spans integer and real-number auxiliaries, digit and
floating-point definitions, generic formats, and
IEEE~754-oriented interfaces. These components form a deep dependency
structure: foundational arithmetic supports generic-format definitions, which
in turn support rounding, precision-specific formats, error results, and
IEEE-facing developments.

This structure makes Flocq a useful repository-scale translation target. High-level modules depend on long chains of earlier declarations, while several foundational interfaces have large downstream reach. Translation therefore requires not only proving individual target theorems, but maintaining compatible representations and interfaces throughout the development.

\subsection{Translation Artifact}

The final translation covers the complete in-scope source inventory. It builds from a clean Lean environment without sorries or other admitted proof obligations. The released translation index records the correspondence between source and target items and is used both for repository construction and evaluation. The source code is available in the supplementary material.

\subsection{Case Analysis: When Compilation Hides Semantic Drift}

We ground the repository-level failure modes above in one representative repair from the \artifact{} history. Commit \texttt{ccf413dc} changed \texttt{new\_location\_odd} in \texttt{Calc/Bracket.lean}, a helper that combines an odd-step index with a local location result. The example is small enough to execute, but it exposes a distinction that is easy to miss when a translation is judged only by target-side compilation.

\paragraph{A concrete semantic mismatch.}
For \texttt{k = 0}, the Flocq definition maps every inexact input location to \texttt{Lt}. The historical Lean translation instead preserved the incoming ordering when \texttt{nb\_steps = 1}. Figure~\ref{fig:bracket-case} makes the difference observable: on \texttt{(nb\_steps, k, l) = (1, 0, Inexact gt)}, the historical target returns \texttt{Inexact gt}, whereas both Flocq and the repaired Lean definition return \texttt{Inexact lt}. Both Lean versions type-check; only the source comparison distinguishes them.

\begin{figure*}[t]
\centering
\begin{minipage}[t]{0.47\textwidth}
\textbf{Flocq source and historical Lean translation}
\begin{lstlisting}[basicstyle=\ttfamily\scriptsize,frame=single,breaklines=true,columns=fullflexible,keepspaces=true,showstringspaces=false]
(* Flocq: k = 0 branch *)
match l with
| loc_Exact => l
| _ => loc_Inexact Lt
end

-- Lean before the repair
| Location.loc_Inexact ord =>
    if nb_steps = 1 then
      Location.loc_Inexact ord
    else
      Location.loc_Inexact Ordering.lt
\end{lstlisting}
\end{minipage}
\hfill
\begin{minipage}[t]{0.47\textwidth}
\textbf{Repaired Lean translation and regression witness}
\begin{lstlisting}[basicstyle=\ttfamily\scriptsize,frame=single,breaklines=true,columns=fullflexible,keepspaces=true,showstringspaces=false]
-- Lean after ccf413dc
| Location.loc_Inexact _ =>
    Location.loc_Inexact Ordering.lt

example :
  new_location_odd 1 0
      (Location.loc_Inexact Ordering.gt) =
    Location.loc_Inexact Ordering.lt := by
  simp [new_location_odd]
\end{lstlisting}
\end{minipage}
\caption{A compiling but semantically divergent translation of \texttt{new\_location\_odd}. The target-only \texttt{nb\_steps = 1} branch changes an observable result; commit \texttt{ccf413dc} removes the branch and adds the shown regression witness.}
\label{fig:bracket-case}
\end{figure*}

\paragraph{Root cause: a theorem premise became program behavior.}
The Coq section assumes \texttt{1 < nb\_steps} for correctness lemmas, but \texttt{new\_location\_odd} itself does not use the hypothesis; Coq therefore omits the premise from the exported definition, which remains callable at \texttt{nb\_steps = 1}. The historical Lean version instead encoded this proof-side restriction as executable behavior. Target compilation cannot reveal that category error.

\paragraph{Why the repair is repository-level.}
The faithful fix does more than delete one branch. The same commit restores an explicit \texttt{Hnb\_steps : 1 < nb\_steps} premise to \texttt{new\_location\_even\_correct}, \texttt{new\_location\_odd\_correct}, and \texttt{new\_location\_correct}; their downstream caller \texttt{inbetween\_float\_new\_location} must then prove the corresponding fact \texttt{1 < p}. Thus the computational interface stays source-faithful while the logical precondition propagates through the dependency graph. A compiler-only repair loop has an incentive to absorb such obligations into a convenient local definition. Source-grounded validation and explicit dependency tracking instead expose the behavioral counterexample, identify the missing theorem premise, and make the downstream proof obligation visible. This single repair therefore illustrates why compilation is a necessary first step but cannot substitute for semantic alignment: translation accuracy requires both.

\section{Experiments}

We evaluate whether dependency-aware repository control improves formal
repository translation beyond item decomposition and compiler-guided repair.
All controlled comparisons use matched source tasks, target environments, model
configurations, and generation budgets.

\subsection{Task and Metric Selection}

\paragraph{Tasks}
We sample 22 dependency-connected submodules from the Flocq translation
task, spanning foundational, generic-format, rounding, calculation, and
IEEE-oriented components. Each task retains its induced source dependency
graph, enabling matched comparisons across different dependency depths and
downstream reach. Section~3 reports the complete 5,075-item \artifact{};
smaller submodules make repeated, matched strategy comparisons tractable.

\paragraph{Metrics}
We evaluate each generated translation in two steps. \textit{Compilation} checks whether the target declaration compiles in the accumulated repository. For translations that compile, \textit{semantic alignment} assesses whether the target declaration agrees with its source counterpart under the fixed source-grounded evaluator. Translations that do not compile do not proceed to semantic evaluation.

We aggregate these two aspects into \textbf{translation accuracy}. For an
evaluation set $V_{\mathrm{eval}}$, translation accuracy is
\[
    \operatorname{Acc}_{\mathrm{trans}}
    = \frac{1}{|V_{\mathrm{eval}}|}
      \sum_{v \in V_{\mathrm{eval}}}
      \mathbf{1}\!\left[C(v)=1 \land A(v)=1\right],
\]
where $C(v)$ indicates successful compilation and $A(v)$ indicates semantic
alignment. A source item therefore contributes to the numerator only when its
translation passes both steps; compilation failures remain in the denominator
and count as unsuccessful translations.

\subsection{Baseline Construction}

All systems share the model, source scope, target toolchain, retry budget, and
compiler interface. \textbf{Flat} translates a submodule as one task;
\textbf{Item-wise} decomposes it without dependency scheduling;
\textbf{Compiler loop} adds bounded target-prover feedback; and
\textbf{\ours{}} adds dependency-ready execution and stable upstream
interfaces to the same item worker and compiler loop. Thus, the primary
Compiler-loop--\ours{} comparison isolates repository coordination while
holding local generation and repair capabilities fixed.

\subsection{Experiment Results}

On the 22 tasks, \ours{} achieves 31.36\% translation accuracy with Claude Opus~5, compared with 12.71\% for the compiler loop under the same model. Across the five models, \ours{} improves translation accuracy despite differences in baseline accuracy. Specifically, \ours{} increases the performance of Gemini 3.1 Pro Preview from 2.54\% with raw LLM prompting to 23.73\%, nearly 10 times its original accuracy, showing that dependency-ordering and freezing techniques are vital to efficient formal repository translation.

\begin{figure*}[t]
  \centering
  \includegraphics[width=\textwidth]{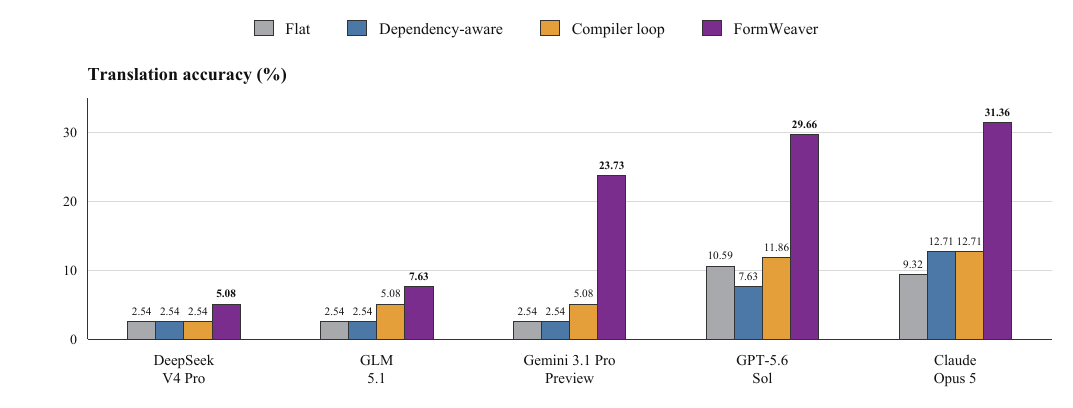}
  \caption{Translation accuracy across five frontier models and four repository
  translation strategies on the matched 118-item evaluation set. A translated
  item counts as accurate only when it first passes the target compilation step
  and then preserves the source semantics under the fixed evaluator; bar
  heights therefore report items that pass both checks as a
  percentage of all 118 source items, with higher values indicating stronger
  translation accuracy. \ours{} achieves the
  highest score for every evaluated model, with the largest absolute
  improvements over the compiler loop on Claude Opus~5 and Gemini~3.1 Pro
  Preview.}
  \label{fig:main-results}
\end{figure*}

\subsection{Ablation: Repository-Control Components}

We ablate the three repository-control mechanisms in \ours{} while retaining
the same item-level worker and compiler loop: dependency scheduling, upstream
interface stabilization, and scoped repair.

\begin{table*}[t]
    \caption{Component ablations of \ours{}. Outcome rates are percentages; Calls/item is the mean number of model calls per item. \ours{} achieves the best performance on compile success rate, semantic faithfulness score, and overall task completion. Not counting the no-repair baseline, \ours{} is also as efficient, or even more efficient, in terms of model calls per item.}
    \label{tab:ablation}
    \centering
    \small
    \resizebox{\textwidth}{!}{%
    \begin{tabular}{lccccccc}
        \toprule
        Variant &
        \shortstack{Dependency\\schedule} &
        \shortstack{Stable\\interfaces} &
        \shortstack{Scoped\\repair} &
        \shortstack{Compile\\success} &
        \shortstack{Semantic\\faithfulness} &
        \shortstack{Task\\completion} &
        Calls/item \\
        \midrule
        \rowcolor{gray!15}
        Full \ours{}            & \checkmark & \checkmark & \checkmark & 30.9\% & 96.6\% & 20.0\% & 2.084 \\
        w/o scheduling          & -- & \checkmark & \checkmark & 22.6\% & 87.7\% & 0.0\% & 3.287 \\
        w/o stabilization      & \checkmark & -- & \checkmark & 23.7\% & 92.2\% & 0.0\% & 2.032 \\
        w/o scoped repair       & \checkmark & \checkmark & -- & 17.6\% & 73.9\% & 0.0\% & 1.000 \\
        \bottomrule
    \end{tabular}%
    }
\end{table*}
        
Full \ours{} attains 30.9\% compile success, 96.6\% semantic faithfulness, and 20.0\% task completion. Removing scheduling or stabilization reduces compile success to 22.6\% and 23.7\%, respectively, and eliminates task completion. Removing scheduling also raises calls per successful item from 2.084 to 3.287, while removing stabilization yields 2.032.
We compare compilation, semantic faithfulness, task completion, and repair
propagation. This separates mechanisms that mainly improve local generation
from those whose effect appears only after translated items are integrated into
a repository.

\subsection{Ablation: Dependency Scaling}

We test how translation success changes as source dependencies deepen using
GPT-5.6 Sol. At each depth $d \in \{0,1,2,3,4\}$, we sample 20 cases and
evaluate the same cases under Raw LLM, dependency-aware scheduling, a compiler
loop, and \ours{}, giving 100 cases per setting across the five depth levels.

\begin{figure}[t]
  \centering
  \includegraphics[width=0.94\linewidth]{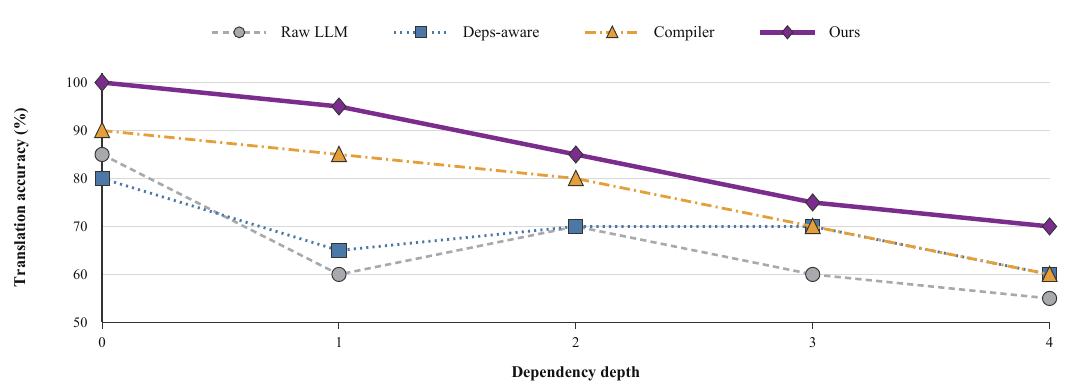}
  \caption{Translation accuracy as source dependencies deepen under GPT-5.6
  Sol. For each depth $d\in\{0,1,2,3,4\}$, all four strategies are evaluated on
  the same 20 cases.
  Points show the fraction of translations that pass both evaluation steps at each depth, with higher values indicating greater robustness. \ours{} leads at every
  depth, consistently outperforming the other baselines.}
  \label{fig:dependency-scaling}
\end{figure}

Figure~\ref{fig:dependency-scaling} shows that \ours{} has the highest mean
success rate, 85\%, compared with 77\% for the compiler loop, 69\% for
dependency-aware scheduling, and 66\% for Raw LLM. The result therefore supports an aggregate benefit from combining repository controls. Both
\ours{} and the compiler loop fall by 30 percentage points from depth 0
to depth 4, and the advantage of \ours{} over the compiler loop varies. Compute and annotation budgets limited the size of this study.

\section{Related Work}

\paragraph{Formal proof and verified code generation}
LLMs have been widely studied for formal theorem proving and verified code generation. LeanDojo~\citep{yang2023leandojo} provides an interactive environment and benchmark for retrieval-augmented theorem proving, while Verina~\citep{ye2026verina} and CLEVER~\citep{thakur2026clever} evaluate generation of implementations and proofs from formal specifications. More recent work has begun to move beyond isolated obligations. VeriSoftBench~\citep{xin2026verisoftbench} evaluates formal reasoning under repository-level context, P$^3$~\citep{li2026p} jointly plans programs and proofs, and Vero~\citep{ye2026vero} studies construction of multi-module formally verified software repositories. These works demonstrate the growing ability of agents to operate over larger formal developments, but generally assume that the target specifications to be implemented or proved are already available. Our setting is complementary: formal repository translation must additionally construct and preserve the target specifications and interfaces themselves while maintaining correspondence to an existing source development.

\paragraph{Autoformalization}
A second line of work translates informal mathematical or software descriptions into formal statements and developments. Early work studied statement-level autoformalization~\citep{wu2022autoformalization,azerbayev2023proofnet}, while recent systems and benchmarks investigate larger theories and domain-specific formal developments~\citep{feng2026theory,ren2026merlean,wang2026m2f}. A central challenge in this setting is evaluating whether the generated formal statement faithfully captures the informal source; recent work therefore develops reference-based and semantic-alignment metrics for autoformalization~\citep{poiroux2025reliable,liu2025rethinking,han2026shadowbench,cornish2026faithformbench}. Formal repository translation differs in that the source is already formal: it provides explicit declarations, dependencies, and machine-checkable semantics that can guide repository construction. The challenge shifts from recovering formal meaning from natural language to preserving formal meaning while adapting representations, interfaces, and proofs across proof ecosystems.

\paragraph{Cross-prover translation}
Cross-prover translation is the closest setting to our work. MiniF2F in Rocq studies automatic transfer of theorem statements and proofs between proof assistants~\citep{viennot2025minif2f}; Babel-formal translates proofs between Lean and Rocq~\citep{stoskopf2025babelformal}; Lean2Isabelle studies factorized proof translation across assistants~\citep{liu2026lean2isabelle}; and ITPEval systematically evaluates formal translation across multiple interactive theorem provers~\citep{wu2026itpeval}. The lf-lean/Rocq-Dove effort further demonstrates strong certificate-based correspondence for supported Lean--Rocq translations~\citep{theorem2026lflean}. These works primarily study whether individual statements, proofs, or supported collections can be translated correctly. In contrast, we focus on the \emph{construction process} required to translate a dependency-rich repository: deciding when declarations are ready to translate, stabilizing upstream representations and interfaces, constraining repair, and maintaining an integrated target repository as translation proceeds. \ours{} therefore complements item-level translation methods with repository-level execution control.

\paragraph{Floating-point formalization}
Our artifact builds on a long line of formalization work for floating-point arithmetic. Flocq~\citep{boldo2011flocq} provides a comprehensive Coq library for reasoning about floating- and fixed-point arithmetic and has supported verified compilation of floating-point computations~\citep{boldo2015verified}. In Lean, FLoPS formalizes semantics and properties of P3109 floating-point representations~\citep{chang2026flops}, while TorchLean develops explicit floating-point semantics for formally verified neural-network execution~\citep{george2026torchlean}. \artifact{}'s contribution is instead the complete translation of an established, dependency-rich Coq repository into Lean, serving both as reusable formal infrastructure and as evidence that repository-scale cross-prover construction is feasible.

\section{Limitations and Conclusion}

\paragraph{Limitations}
Our study focuses primarily on one substantial source-target pair, Coq-to-Lean translation, and one domain, floating-point formalization. Although Flocq provides a large and dependency-rich test case, the relative importance of dependency scheduling, interface stabilization, and scoped repair may differ for other proof assistants and domains with different module systems, libraries, or proof engineering conventions. Formal repository translation also assumes the existence of a trusted formal source and therefore does not address settings in which specifications must be recovered directly from informal descriptions. In addition, source-target semantic correspondence is not always reducible to target-side compilation: representation bridges and library-level semantic differences may require explicit correspondence definitions or additional validation. Finally, repository-scale translation remains computationally expensive, and some architectural choices, such as target representations or library interfaces, may still require careful design even when subsequent proof and implementation construction is largely automated.

\paragraph{Conclusion}
We studied formal repository translation as a means of scaling LLM-assisted formal development beyond isolated declarations while reusing existing formalization effort across proof ecosystems. We introduced \ours{}, a dependency-aware translation pipeline that coordinates item-level generation through explicit dependency scheduling, stabilized upstream interfaces, and scoped compiler-guided repair. Using \ours{}, we construct \artifact{}, a complete Lean translation of Flocq comprising 58 core files, 5,075 indexed items, and 194,704 lines of Lean code. Controlled experiments further show that repository-level coordination substantially improves translation accuracy, requiring both successful compilation and semantic alignment, over simpler translation strategies. The central challenge is repository-level coherence: specifications, representations, interfaces, and proofs must remain compatible as the target development grows.

\begin{ack}
This work was carried out in collaboration with the Beneficial AI Foundation. We thank Max Tegmark for the guidance provided throughout this project.
\end{ack}

\bibliographystyle{plainnat}
\bibliography{references}

\clearpage
\appendix
\input{appendix.tex}

\newpage
\input{checklist.tex}

\end{document}

%% file: appendix.tex
\section{Model and Prompt Setup}
\label{app:model-prompt-setup}

This appendix summarizes the model-facing configuration shared by the
controlled translation runs. The five evaluated model families are named in
Figure~\ref{fig:main-results}. All translation strategies use the same agent
runtime and target Lean environment within a matched model condition; the
strategies differ in how work is decomposed, scheduled, and repaired, as
described in Section~4.

\subsection{Agent runtime}

Each invocation launches one agent attempt in the repository working directory
and records the selected model and reasoning profile. The runner does not
override temperature, top-$p$, top-$k$, sampling seed, context-window size, or
maximum output length, so these remain client or provider defaults and we do
not claim deterministic sampling. The agent has repository-scoped filesystem
and shell access, including the Lean compiler and build system. Each attempt has
a 1,800-second timeout; at most one retry is allowed for a recognized
transport-level failure, while translation failures are not reclassified as
infrastructure failures. The runner retains the prompt, event stream, final
response, code changes, compiler output, and build result.

\subsection{Prompt construction and guardrails}

The task prompt is assembled from four layers: repository-wide agent rules, the
source-faithful translation protocol, a strategy-specific task template, and
runtime fields such as the assigned target and current compiler diagnostics.
Across strategies, the shared instructions require the agent to:

\begin{enumerate}
  \item operate only on the assigned translation target and avoid unrelated
  repository edits;
  \item consult the upstream Flocq declaration when changing a definition or
  theorem statement;
  \item avoid admitted proofs, new axioms, tautological specifications,
  identity stubs, and patches that assume the desired conclusion;
  \item report a blocker instead of weakening the source semantics when a
  required foundational result is unavailable; and
  \item run the Lean build after changing code and report whether the attempt
  was proved, blocked, failed, or required no action.
\end{enumerate}

These shared constraints keep local generation capabilities fixed across the
matched comparison. The baseline-specific differences described in
Section~4.2 concern repository control: flat or item-level decomposition,
dependency-ready scheduling, compiler-guided repair, and stabilization of
accepted upstream interfaces.

\subsection{Compiler feedback and evaluation separation}

During generation, the agent can inspect repository files and obtain Lean
diagnostics by invoking the target toolchain. Compiler messages may therefore
guide bounded repair in strategies that include a compiler loop. The agent does
not receive semantic-alignment verdicts or held-out evaluator labels. Semantic
alignment is assessed only after a translation passes the compilation step, as
defined in Section~4.1. This separation prevents evaluator feedback from being
used as an additional repair signal and preserves the two-stage translation-
accuracy protocol used in the paper.

%% file: checklist.tex
\section*{NeurIPS Paper Checklist}

\begin{enumerate}

\item {\bf Claims}
    \item[] Question: Do the main claims made in the abstract and introduction accurately reflect the paper's contributions and scope?
    \item[] Answer: \answerYes{}
    \item[] Justification: The abstract and introduction state the repository-translation problem, the dependency-aware pipeline, the scale of the Flocq-to-Lean artifact, and the comparative translation-accuracy claim reported in the controlled experiments. The paper also limits its empirical scope to one source--target pair and one formalization domain.

\item {\bf Limitations}
    \item[] Question: Does the paper discuss the limitations of the work performed by the authors?
    \item[] Answer: \answerYes{}
    \item[] Justification: The dedicated Limitations paragraph discusses the single Coq-to-Lean source--target pair, the floating-point domain, reliance on a trusted formal source, representation-dependent semantic correspondence, computational expense, and the remaining need for architectural judgment.

\item {\bf Theory assumptions and proofs}
    \item[] Question: For each theoretical result, does the paper provide the full set of assumptions and a complete (and correct) proof?
    \item[] Answer: \answerNA{}
    \item[] Justification: The paper presents an empirical repository-translation system and evaluation design, with no new mathematical theorem or theoretical convergence claim.

    \item {\bf Experimental result reproducibility}
    \item[] Question: Does the paper fully disclose all the information needed to reproduce the main experimental results of the paper to the extent that it affects the main claims and/or conclusions of the paper (regardless of whether the code and data are provided or not)?
    \item[] Answer: \answerYes{}
    \item[] Justification: The paper specifies the 22 dependency-connected tasks, four translation strategies, five-model comparison, sequential compilation and semantic-alignment evaluation, matched controls, ablations, and aggregate calculations. Appendix~\ref{app:model-prompt-setup} further documents the shared agent runtime, prompt construction, and feedback available to the model. 

\item {\bf Open access to data and code}
    \item[] Question: Does the paper provide open access to the data and code, with sufficient instructions to faithfully reproduce the main experimental results, as described in supplemental material?
    \item[] Answer: \answerYes{}
    \item[] Justification: The current submission provides an anonymized supplementary containing the translation pipeline, full artifact of FloatSpec repo, and analysis scripts.

\item {\bf Experimental setting/details}
    \item[] Question: Does the paper specify all the training and test details (e.g., data splits, hyperparameters, how they were chosen, type of optimizer) necessary to understand the results?
    \item[] Answer: \answerYes{}
    \item[] Justification: The paper identifies the task family, sampling unit, evaluated strategies and models, matched controls, sequential metric, and ablation designs. Appendix~\ref{app:model-prompt-setup} describes the shared runner, prompt layers, decoding defaults, timeout, tool access, and retry policy. 

\item {\bf Experiment statistical significance}
    \item[] Question: Does the paper report error bars suitably and correctly defined or other appropriate information about the statistical significance of the experiments?
    \item[] Answer: \answerYes{}
    \item[] Justification: For the comparisons supporting the main claim, the paper reports 95\% Wilson confidence intervals for compilation success, strict semantic faithfulness, and repository-complete performance, and identifies the corresponding numerators and denominators. The paper also notes when intervals overlap and avoids interpreting the small dependency-depth sample as a precise monotonic trend.

\item {\bf Experiments compute resources}
    \item[] Question: For each experiment, does the paper provide sufficient information on the computer resources (type of compute workers, memory, time of execution) needed to reproduce the experiments?
    \item[] Answer: \answerNo{}
    \item[] Justification: The paper reports measured agent-call counts for the main comparison, including 745 calls over 66 artifact-backed \ours{} cells and 155 calls over 110 compiler-loop cells, and explicitly notes that the former is a lower-bound cost comparison. It does not yet report hardware, memory, wall time, token usage, or total project compute.
    
\item {\bf Code of ethics}
    \item[] Question: Does the research conducted in the paper conform, in every respect, with the NeurIPS Code of Ethics \url{https://neurips.cc/public/EthicsGuidelines}?
    \item[] Answer: \answerYes{}
    \item[] Justification: The work uses formal-code artifacts, involves no personal data or human subjects, and distinguishes compilation, finite semantic evidence, and proof certificates to reduce the risk of overstating correctness.

\item {\bf Broader impacts}
    \item[] Question: Does the paper discuss both potential positive societal impacts and negative societal impacts of the work performed?
    \item[] Answer: \answerNA{}
    \item[] Justification: This work is not targeted to social aspects and is focusing on auto-formalization.
    
\item {\bf Safeguards}
    \item[] Question: Does the paper describe safeguards that have been put in place for responsible release of data or models that have a high risk for misuse (e.g., pre-trained language models, image generators, or scraped datasets)?
    \item[] Answer: \answerNA{}
    \item[] Justification: The work introduces neither a general-purpose model nor scraped personal or media data; the formal-code artifacts do not create a distinct high-misuse-risk release.

\item {\bf Licenses for existing assets}
    \item[] Question: Are the creators or original owners of assets (e.g., code, data, models), used in the paper, properly credited and are the license and terms of use explicitly mentioned and properly respected?
    \item[] Answer: \answerYes{}
    \item[] Justification: Flocq and the relevant proof-assistant and comparison-system work are cited, and the paper provided the redistribution terms for the source material and generated target artifact in the supplememtary.

\item {\bf New assets}
    \item[] Question: Are new assets introduced in the paper well documented and is the documentation provided alongside the assets?
    \item[] Answer: \answerYes{}
    \item[] Justification: The paper documents the scope and architecture of FloatSpec and reports its size, and it provides an anonymized release and a license in the supplementary.

\item {\bf Crowdsourcing and research with human subjects}
    \item[] Question: For crowdsourcing experiments and research with human subjects, does the paper include the full text of instructions given to participants and screenshots, if applicable, as well as details about compensation (if any)? 
    \item[] Answer: \answerNA{}
    \item[] Justification: The research does not involve crowdsourcing or human-subject experiments; expert artifact review was ordinary research work by the authors.

\item {\bf Institutional review board (IRB) approvals or equivalent for research with human subjects}
    \item[] Question: Does the paper describe potential risks incurred by study participants, whether such risks were disclosed to the subjects, and whether Institutional Review Board (IRB) approvals (or an equivalent approval/review based on the requirements of your country or institution) were obtained?
    \item[] Answer: \answerNA{}
    \item[] Justification: The research does not involve human subjects or collection of personal data, so IRB review is not applicable.

\item {\bf Declaration of LLM usage}
    \item[] Question: Does the paper describe the usage of LLMs if it is an important, original, or non-standard component of the core methods in this research? Note that if the LLM is used only for writing, editing, or formatting purposes and does \emph{not} impact the core methodology, scientific rigor, or originality of the research, declaration is not required.
    %this research? 
    \item[] Answer: \answerYes{}
    \item[] Justification: The method and experimental-design sections describe the agent's item-level translation and bounded-repair roles and distinguish them from deterministic indexing, scheduling, compilation, and semantic replay. Appendix~\ref{app:model-prompt-setup} additionally summarizes the model-facing runtime, prompt construction, guardrails, and separation from semantic-evaluator labels.

\end{enumerate}